# Exploring the Use of a Large Language Model for Data Extraction in Systematic Reviews: a Rapid Feasibility Study

Lena Schmidt[1,*], Kaitlyn Hair[2], Sergio Graziosi[3], Fiona Campbell[1], Claudia Kapp[4], Alireza Khanteymoori[5], Dawn Craig[1], Mark Engelbert[6,7] and James Thomas[3]

[1] *National Institute for Health and Care Research Innovation Observatory, Population Health Sciences Institute, Newcastle University, Newcastle, UK*
[2] *Centre for Clinical Brain Sciences, University of Edinburgh, UK*
[3] *UCL Social Research Institute, University College London, London, UK*
[4] *Institute for Quality and Efficiency in Health Care, Cologne, Germany*
[5] *Department of Neurosurgery, Neurocenter, Medical Center - University of Freiburg, Freiburg, Germany*
[6] *International Initiative for Impact Evaluation (3ie)*
[7] *School of International Development, University of East Anglia, Norwich, UK*

**Abstract**
This paper describes a rapid feasibility study of using GPT-4, a large language model (LLM), to (semi)automate data extraction in systematic reviews. Despite the recent surge of interest in LLMs there is still a lack of understanding of how to design LLM-based automation tools and how to robustly evaluate their performance.

During the 2023 Evidence Synthesis Hackathon we conducted two feasibility studies. Firstly, to automatically extract study characteristics from human clinical, animal, and social science domain studies. We used two studies from each category for prompt-development; and ten for evaluation. Secondly, we used the LLM to predict Participants, Interventions, Controls and Outcomes (PICOs) labelled within 100 abstracts in the EBM-NLP dataset.

Overall, results indicated an accuracy of around 80%, with some variability between domains (82% for human clinical, 80% for animal, and 72% for studies of human social sciences). Causal inference methods and study design were the data extraction items with the most errors. In the PICO study, participants and intervention/control showed high accuracy (>80%), outcomes were more challenging. Evaluation was done manually; scoring methods such as BLEU and ROUGE showed limited value. We observed variability in the LLMs predictions and changes in response quality.

This paper presents a template for future evaluations of LLMs in the context of data extraction for systematic review automation. Our results show that there might be value in using LLMs, for example as second or third reviewers. However, caution is advised when integrating models such as GPT-4 into tools. Further research on stability and reliability in practical settings is warranted for each type of data that is processed by the LLM.

**Keywords**
Large Language Models, Automation, Systematic Reviews, Automation of Systematic Reviews, Reproducibility, Reliability of AI

## 1. Introduction

The use of automation is established practice in many systematic reviews and other types of evidence synthesis and has been used across the review process from search strategy development, helping to screen records for eligibility and to support data extraction and risk of bias assessment.[1] However, the use of automation has been fairly limited, and most reviews follow a mainly manual workflow. This may be about to change. In large part due to the widespread use of ChatGPT, there is increasing interest in using large language models (LLMs) to support systematic reviews, with new 'AI' systematic review tools appearing ever more frequently.

In contrast to more conventional approaches to machine learning, where tools have a specified purpose (e.g. the classification of research design in health), the same LLM might be able to assist with multiple tasks in the same review, including screening, data extraction, risk of bias detection, and (most controversially) synthesis. Thus, it may be that LLMs are a more disruptive technology to established systematic review practices than more conventional machine learning has hitherto been. There are now many tools that claim to be able to make reviews more efficient using language models, but which lack any robust evaluation.

There is thus an urgent need for the systematic review community to understand the strengths and limitations of these new LLM-based tools. The very versatility of ChatGPT and similar LLMs makes them a challenging target for evaluation, as they are designed to be 'general' language models, and it is believed they can be used in a wide range of tasks. In addition, they are 'black boxes' – unable to explain why a given output was generated, and often, unable to state how likely it is to be correct.

This paper is an attempt to begin to address the above issues. We are not claiming that it is the definitive evaluation of the use of LLMs for data extraction in systematic reviews but do report the result of an evaluation of the use of GPT-4 for this zero-shot classification task. It is the result of three days of intensive work in Newcastle at the Evidence Synthesis Hackathon.[1] As well as tempering some of the extravagant claims circulating about the amazing capabilities of LLMs, it aims to provide a template for further evaluations, emphasising transparency in reporting and the careful separation of datasets used for prompt design and those used for evaluation.

There are several reasons for this paper to focus on data extraction, rather than screening, which is well known to be extremely time consuming and somewhat amenable to automation. First, asking the machine "what is described here?" defines a task that the machine can potentially perform accurately, while asking "can the study described here help to answer this question?" defines a highly intellectual task that LLMs are not designed to perform.

A second reason concerns the cost of undetected mistakes: erroneously excluding relevant evidence is an error associated with the highest cost in our field, and given the nature of screening, an error that may well go undetected. On the other hand, making the occasional mistake in extracting atomic snippets of information has an inherently lower cost, as well as a higher probability of being detected during related synthesis steps.

---

[1] https://www.eshackathon.org/

## 2. Methods

### 2.1. Research questions

1. How does the extraction of data items typical of those extracted in systematic reviews compare between a LLM and humans?
2. Does performance differ according to domain of question or domain of research?
3. How stable are the results from the LLM? (i.e. does repeated prompting yield the same responses?)

### 2.2. Study design

On the first day of the hackathon, we experimented with writing prompts to extract data and designed two evaluative feasibility studies to evaluate using a LLM for data extraction: a 'comparative' study and a 'PICO extraction' study.

1. In the comparative study, we assessed the ability of a LLM to extract data from abstracts of research reports in three domains: human clinical trials; social science evaluations; and animal studies. This study addressed all three research questions.
2. In the PICO extraction study, we used the LLM to predict Patient, Intervention/Control, and Outcome (PICO) entities from abstracts of the EBM-NLP dataset.[2] This study addressed research question 1 only.

## 3. Data

### 3.1. Comparative study

Because we wanted to see how performance of the LLM varied by research domain, we decided to focus on three domains: human social science, human clinical, and animal research. Bearing in mind that this was a rapid study that was conducted intensively at a hackathon, we decided that we had capacity to process 36 studies manually.

Twelve studies in the "human social science" group were retrieved from the Development Evidence Portal, maintained by International Initiative for Impact Evaluation (3ie), which contains impact evaluations conducted in low- and middle-income countries. The portal's "Sector" filter was used to select studies from several different sub-disciplines, including education, agricultural economics, and public health.

We obtained six animal studies from PubMed using the basic search functionality (search string: alzheimer AND mice AND "open field"). This was to ensure that we collected animal studies that were likely to report outcome measures (the open field test is one of the most commonly reported tests in the animal literature). The other six animal studies were chosen from a pool of pre-screened papers. These pre-screened papers were specifically related to animal studies in spinal cord injury. To enhance the diversity of the dataset, papers with different outcomes were incorporated.

The 12 studies in the "human clinical studies" group were identified through a PubMed search for the publication type "randomized controlled trial", searching for "behavioural intervention" (free text) and filtering to studies published after 2015. The aim was to identify studies which might have reasonable reporting, but also avoiding simple drug treatment evaluations.

In each of the three domains we split the data into train and test sets, with two studies in the 'train' set and 10 in the 'test' set. The three pairs of studies in the 'train' set were used for prompt development (below), while the ten studies in each test set were held out for testing the automated data extraction against. A 'gold standard' data extraction was generated for each by one of six members of the team. We did not look at the test papers until prompt development was completed; this ensured that we did not inadvertently contaminate the evaluation with prior 'knowledge' of the test set.

### 3.2. PICO study

For the PICO study, we selected 100 studies from the EBM-NLP dataset [2]. This dataset contains the titles and abstracts from 5000 human clinical trials. We limited our study to 100 titles and abstracts because we evaluated predictions manually. Text spans which represent the Population, Intervention, Comparator, and Outcome have been extracted by a diverse range of expert and non-expert human 'workers'. The aim of the dataset is to support research and development of natural language processing tasks concerning study PICO, so it is a good match for our own evaluation. For this purpose, one author manually compared EBM-NLP labels with LLM predictions. Such manual evaluations are more time-consuming than automatic evaluations in the form of precision, recall, and F1 scores that are traditionally used for information extraction tasks [2]. The generative nature of LLM output in our study means that these traditional scoring methods are of limited utility. To explore alternatives, we computed BLEU[2] and ROUGE[3] scores between gold-standard labels and the LLM answers. These methods evaluate the similarity between gold-standards and predictions, for example through word overlap. We chose them because they are frequently used to evaluate summarisation and translation tasks, and might therefore be better fitting than recall or F1 scores[2]. The evaluation script is available via our GitHub repository.[4]

## 4. Prompt development

### 4.1. Comparative study

Prior to the hackathon, we deployed a new feature into EPPI Reviewer which enabled prompts to be created for codes in a coding tool; for the prompts and the abstract of a study to be submitted to the GPT-4 API; and for results to be returned and appraised.

---

[2] https://www.nltk.org/api/nltk.translate.bleu_score.html
[3] https://pypi.org/project/rouge-score/
[4] https://github.com/L-ENA/ES-hackathon-GPT-evaluation

We took the two studies in the 'train' set in each domain and composed prompts to extract data under the following headings: study design, subjects, study on humans, study on animals, N of subjects in study, comparisons, outcomes measured. For each heading, one team member wrote an initial prompt to ensure all domains would develop prompts from the same starting point. For human social science studies four additional areas were added: number of arms, causal inference method[5], country, and intervention description.

We found that prompts were sensitive to:

- **Minor changes in wording**. For example, adding 'complete' to the prompt when requesting a description of outcomes made a significant difference at times.
- **Changes in position in the sequence of prompts**. One prompt about the details of study participants would yield a good result when placed high in the list, but would often not yield any results at all if placed further down or in isolation without other prompts. This is due to the specific implementation details of the system we used: all prompts are embedded in a single request sent to the GPT API, see below for further details.
- **Changes in the 'field label'** that was given to each prompt in the JSON output. For example, we found that 'n_participants' was a much better label than 'study_size'.

**Appendix B** contains a detailed description of the iterative prompt development process for animal studies.

### 4.2. PICO study

For the PICO study we followed the same approach as above, submitting identical prompts for each of the 100 titles and abstracts:

```
systemPrompt = 'You extract PICO data on clinical trials from the text provided below into a JSON object of the shape provided below. If the data is not in the text return false for that field. \nShape: {population: string // state full details of the population in the study, intervention(s): string // state full details of the interventions in each group, outcome(s): string // state all outcomes reported by the trial}'
```

## 5. Prompt submission to GPT-4

When we were satisfied that we could not further improve the prompts, we applied the prompts to the 30 records in the test set and stored the results in EPPI Reviewer. Data were also manually extracted on the 30 studies in each of the domains and stored in the same software.

GPT-4 prompts are structured in a 'system' and 'user' way, whereby the 'system' prompt can be used to initialise the model's behaviour. In our case, we wanted to orientate the

---

[5] The 'causal inference method' prompt was added to capture the quasi-experimental analysis methods that are often used to establish causality in non-randomized effectiveness studies in the social sciences.

system to information extraction and to returning results in a JSON structure for ingestion into EPPI Reviewer. The initial prompt was therefore structured like this:

{role = "system", content = "You extract data from the text provided below into a JSON object of the shape provided below. If the data is not in the text return 'false' for that field. \nShape: {" + prompt + "}"}

The 'prompt' variable for each field to be extracted was configurable by users and was composed of items in a list as detailed in **Appendix C**. Thus, all prompts for a given study were submitted in the same script along with the text.

**Appendix D** shows one example of the entire JSON submitted to GPT-4 for one paper. The parameters selected for GPT-4 were designed to be as conservative as possible, aiming for maximum repeatability across repeated requests. They were: temperature=0, frequency_penalty=0, presence_penalty=0, top_p=0.95. The model used was '2023-07-01-preview', accessed via the Azure OpenAI API on 13-15 December 2023.

## 6. Evaluation of GPT-4 accuracy

### 6.1. Comparative study

Two reviewers compared each of the responses provided by GPT-4 with the data we had extracted manually. Each of the model's responses was rated either *complete*, *partial*, or *incorrect*. If the model's response contained all the essential information requested by the prompt, or if the model correctly did not provide a response when the requested information was absent from the abstract, the response was rated *complete*. If the model's response also contained additional information irrelevant to the prompt, it was still rated as *complete*, unless the additional information lessened the overall accuracy of the response, in which case it could be rated as *partial* or *incorrect*.

If the response contained some relevant information, but was missing other essential information, it was rated *partial*. If the model produced an entirely incorrect, wrong, or misleading response, or if it failed to provide a response when the requested information was present in the abstract, it was rated *incorrect*. For example, if the list of outcomes extracted by the model contained several correct responses but also included moderator variables named in the abstract, this was rated *partial.*

Results are presented as descriptive statistics in terms of percentages of responses that fell into each category. Since this is an exploratory study, we did not aim to assess the wider meaning of these scores. However, while humans do not necessarily succeed in attaining 100% accuracy, experience in previous work suggests that systematic review automation tools might be expected to achieve 98 or 99% accuracy [3].

### 6.2. PICO study

One author manually rated results for the first 100 titles and abstracts. The evaluation process was similar to the process used for the 38 main papers described above. However, the evaluating author had access to the gold-standard labels provided within the EBM-NLP corpus, which increased evaluation speed.

# 7. Results

## 7.1. Comparative study (Research Questions 1-3)

Overall, 260 pieces of information were extracted from the 30 studies in the test set. 78% (n=194) were adjudged to be completely correct; 13% (n=32) partially correct; and 10% (n=24) incorrect. There was considerable variation in terms of the type of information being extracted and which domain the study in question was from.

Table 2 summarises accuracy across each type of information being extracted. Higher levels of accuracy are associated with simpler and smaller types of data. For example, whether or not a study involved human or animal subjects; the number of subjects in the study, and the country in which it took place. The language model found it particularly difficult to name the type of study being reported, though there was considerable variation on performance across the three domains of study in response to this question.

**Table 2: overall accuracy across each type of data extracted**

|  | Complete | Partial | Incorrect | Total N papers |
|---|---|---|---|---|
| **Study design** | 47% | 27% | 27% | 30 |
| **Number of arms** | 80% | 0% | 20% | 10 |
| **Causal inference method** | 60% | 0% | 40% | 10 |
| **Country** | 90% | 0% | 10% | 10 |
| **Subjects** | 77% | 20% | 3% | 30 |
| **Study on humans** | 90% | 0% | 10% | 30 |
| **Study on animals** | 100% | 0% | 0% | 30 |
| **N of subject in study** | 90% | 3% | 7% | 30 |
| **Intervention description** | 70% | 30% | 0% | 10 |
| **Comparisons** | 77% | 17% | 7% | 30 |
| **Outcomes measured** | 67% | 30% | 3% | 30 |

Figure 1 summarises the results for the animal studies. Accuracy overall in this domain (80%) was slightly higher than the combined group above. Almost all the incorrect data come from extracting the study design of the research reported in the abstract, and average accuracy would have been higher but for this dimension. Discussion across the group suggested that this is often simply very difficult to assess from the abstract alone. In one instance, the LLM "hallucinated" the study design, indicating that the abstract described a "Randomized controlled trial", when there was no mention of random allocation.

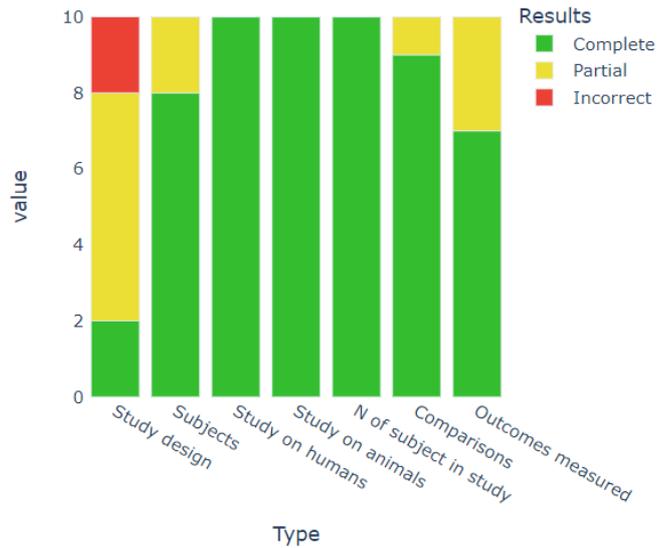

**Figure 1:** results on animal studies (N=10)

Figure 2 reports the results on human clinical studies. This domain had the highest accuracy of the three at 82%. As all abstracts were reports of randomized clinical trials, this was the most homogenous domain, and the automated extraction was fairly accurate across all areas.

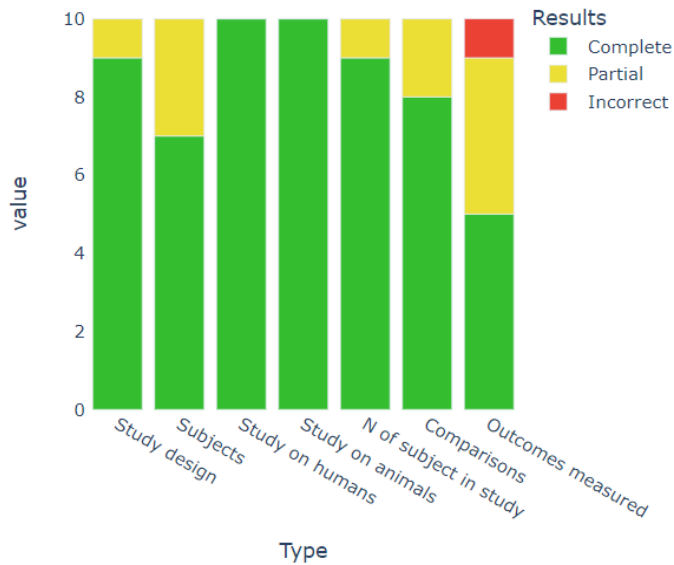

**Figure 2:** results of extraction of human clinical studies (N=10)

Figure 3 reports results for the human social science studies. At 72% accurate, this was the poorest performing domain and the domain where more answers were assessed to be completely, as opposed to partially, incorrect.

This was also the domain where three additional areas of data extraction were tested. The LLM did better at identifying the method for assessing causal inference than it did at identifying the study design. In no areas, and in no studies, did the automated data extraction reach 100%. However, the highest accuracy was for the extraction of country, and this was correctly extracted and mapped against the ISO Alpha-3 code 90% of the time.

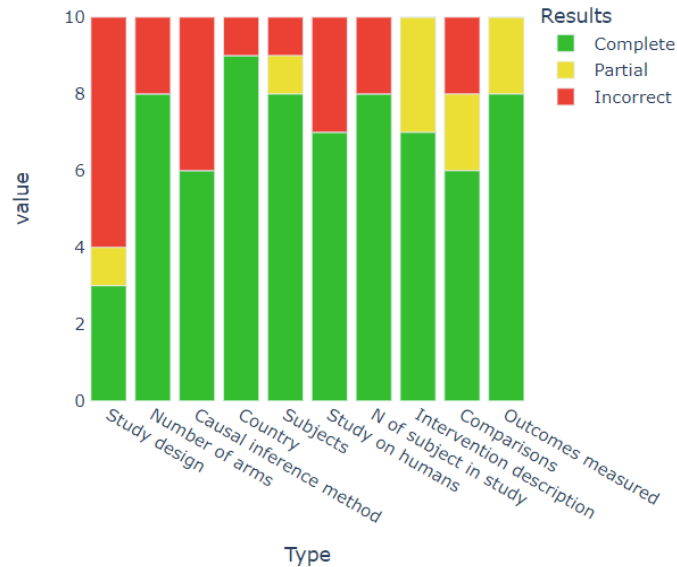

**Figure 3:** results of extraction of human social science studies (N=10)

**Appendix E** contains all accuracy judgements made on the test dataset, while **Appendix F** is the bar-chart representation of the same data.

### 7.1.1. Analysis of response stability

During the prompt development phase, we noticed that sometimes the values produced by GPT-4 changed even when submitting identical prompts. To evaluate the potential impact of such variability (even though for all requests the "temperature" parameter was set to zero, to minimise variation in response) we repeated the automated data extraction a second time against the test set and then evaluated the differences compared to the first round. We found that responses for "Boolean" and "Number" questions were entirely stable. However, for "String" questions, answers were identical only 69% of the time, with small differences 23% of the time and substantive differences 7% of the time. **Appendix G** shows the full results of this analysis.

### 7.2. PICO study (Research Question 1)

The results shown in Figure 4 show a similar trend to the human clinical studies in figure 2. The quality of automatic participant and intervention/control entity extraction was high, with over 80% rated as 'Complete'. As also found above (see Figure 2), outcomes were more challenging to extract completely and accurately.

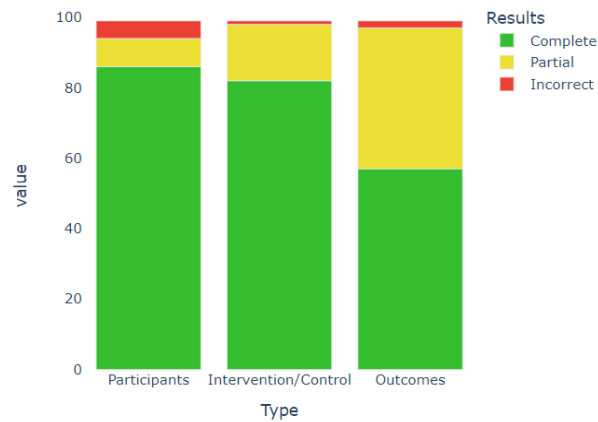

**Figure 4:** results of extraction of human trials within EBM-NLP (N=100)

Figure 4 is based on a manual review of the LLM predictions, carried out by one author. We also computed BLEU[6] and ROUGE[7] scores between gold-standard labels and the LLM answers. These scores are commonly used to evaluate automatic summarisation and translation methods, but their validity for new evaluation scenarios needs to be tested for each new scenario [4]. In our PICO study we found that their results were not meaningful when compared to our human assessment results (data not shown, see appendix). More research is needed to determine data format and evaluation methods for LLM output that uses a gold-standard previously created by humans, rather than a full manual review of results.

## 8. Discussion

### 8.1. Summary of main findings

#### 8.1.1. Research question 1: How does the extraction of data items typical of those extracted in systematic reviews compare between a LLM and humans?

We found that data can be extracted from study abstracts in three domains with 80% overall accuracy and verified this finding on an additional corpus with 100 studies. In a recent living review of automatic data extraction, 76 publications about extracting study characteristics in the form of entities and sentences were included [5]. Typically, entity extraction requires the correct identification of every single occurrence of an entity within the text, which then allows researchers to compute precision, recall, and F1 scores. The task of identifying every entity is different (and potentially harder) than the LLM task, which generates a single unified answer [5]. On the EBM-NLP dataset, where we reached 80% accuracy on 100 abstracts, F1 scores of 76-78% have been described in the literature [6,7]. The caveat, however, remains that the difference between evaluation approaches limits their comparability.

---

[6] https://www.nltk.org/api/nltk.translate.bleu_score.html
[7] https://pypi.org/project/rouge-score/

### 8.1.2. Research question 2: Does performance differ according to domain of question or domain of research?

The machine was most accurate in the human clinical studies domain. The area where the LLM had most difficulty in extracting data accurately was for research in human social science.

The automated data extraction was most accurate when identifying whether a study is conducted on animals, and least accurate when identifying the design of the study. This may be partially because the necessary information is not always reported in the abstract, and because study design is a contested concept with definitions differing across disciplines. Binary and Boolean data types seem to be extracted better than more open response types (strings).

We found that prompts do not necessarily 'travel' well between domains and need to be developed and tested for each use case. As is often the case, performance in social science research lagged the clinical, possibly because of the higher degree of conceptual complexity in the social science domain and the lesser standardisation and structure in social science abstracts.

### 8.1.3. Research question 3: How stable are the results from the LLM? (i.e. does repeated prompting yield the same responses?)

We found that prompt responses were quite consistent between 'runs' for extracting Boolean and numeric data types, but that the extraction of text 'strings' could vary substantively in about 7%, and by a small amount in 23%, of cases.

### 8.2. Strengths and limitations of this study

Despite the study's limited size, it is systematic in its approach, and a major strength is that it considers performance across multiple domains. Its design is robust, ensuring that the data used for testing was not used previously for developing prompts.

It was, of course, carried out rapidly in three days at a hackathon. While this intensive work is a considerable strength, it was also necessary to cut certain corners to complete the work in the time available. Most notably, we did not attempt to create our 'gold standard' data through doing independent double data extraction, and nor did we ensure that the pairs were applying the assessment criteria in the same way; this may have introduced potential bias. (Though the team was all present in the same room, and so talked through decisions regularly.) We also applied the model to abstracts only, rather than full reports, which may limit its generalisability to real-world data extraction scenarios. While we separated training and test data for prompt development and evaluation, it is quite possible that the LLM itself was trained on at least parts of the evaluated data. While this is unlikely to have introduced any substantial bias, it can only be avoided by evaluating research that was published after the training period of the applied LLM.

### 8.3. Implications for use in systematic reviews

Our test has several implications for the use of large language models for data extraction in systematic reviews. Perhaps most importantly, our results suggest that a great deal of caution is needed before deploying LLMs for this purpose. It is probably best to think of LLMs as tools that can semi-automate or serve as a second reviewer on certain tasks, rather than as a way to fully automate all or even part of the data extraction required for a systematic review.

Moreover, because the performance of the model was highly variable across domains and across data types, reviewers should perform detailed testing of LLM performance on each type of data to be extracted. Relatedly, when assessing and reporting model performance, reviewers should pay attention to accuracy for each item individually, rather than focusing only on overall/average accuracy.

Our experience suggests that prompts must be developed and tested iteratively on multiple studies. The first versions of the prompts we used often returned unexpected or unhelpful results. We also found that the ordering of the prompts in a sequence made a difference to the responses. This may have been avoided by submitting each request to GPT-4 in isolation rather than in tandem, but this may also have reduced the overall accuracy of the responses by disallowing the model from relying on the context of previous prompts about a given study. Further testing is needed to determine whether submitting prompts individually or as a group yields better results overall. For any workflow employing the latter approach, reviewers should be mindful of and test for any ordering effects.

Responses to identical prompts about a given paper can differ even when the 'temperature' parameter is set to 0. In other words, the utilization of GPT-4 for data extraction is probably not fully replicable, even though replicability is a hallmark of the systematic review process. Relatedly, the process has limited "explainability", meaning that it can be difficult to predict how an LLM model will respond to a particular prompt about a given paper – and to understand why it did respond in a particular way after the fact. This further underscores the importance of testing and, given that the LLM's output cannot be fully explainable, providing readers with evidence that the model is reliably behaving in the expected manner.

In particular, we noticed that we needed to be careful to ensure that prompts were well-matched to the studies that they were used on. For example, the LLM produced nonsensical results when asked to give details of the intervention in a paper where no intervention was present. We would therefore be cautious about relying on a LLM to make screening decisions, as the set of studies retrieved in a review's search are often highly variable, and the LLM might exclude studies that are relevant but 'look' different to other relevant studies.

### 8.4. Template for future evaluation

Part of our motivation for conducting this study was to create a template that can be used for future evaluation in this area. There are several characteristics of this study which, while small, we feel should be replicated in future work.

First, our results establish that tools using LLMs need to be evaluated before use in real-world reviews. There are very many tools being published without any evaluation or with insufficient evaluation depth. While we support tool development and deployment, we are of the opinion that the use of LLMs in systematic reviews should, at present, be for evaluation only. A challenging aspect of LLM evaluation, limiting LLM utility in practice, is the unstable phrasing of responses. Additionally, the AI's summarised output complicates large-scale quantitative evaluation in terms of sensitivity, precision, and recall, as it is typically performed for algorithms that automate data extraction [2].

Second, as the effectiveness of LLMs is context dependent. LLMs cannot simply be 'dropped into' new reviews and expected to perform in the same way as they may have in other reviews. Prompts need to be checked, fine-tuned and tested for robust performance before use.

Third, it is critical to ensure there is a clear separation of train and test data. While this principle is well established in some fields, there are many situations where researchers are both developing and evaluating prompts on the same data. This mistake is simple to avoid, but avoiding it needs to become expected and standardised.

Fourth, we have been as transparent as we can be with regards to the prompts used, the parameter settings used in the language model, and how the prompts were developed. Other evaluations regularly provide incomplete information on their prompts which renders their results impossible to use or replicate.

This collective effort will not only enhance the credibility of evaluations but also contribute to the responsible and effective integration of LLMs in evidence synthesis. **Appendix H** summarises our initial thoughts about the most important aspects of evaluating LLMs for evidence synthesis and how we have attempted to address them in this study. We invite others to build on this.

## 9. Conclusions

We have found that it is feasible to use a LLM to extract data for use in systematic reviews, though accuracy is limited. This evaluation is small and there is as yet an insufficient evidence base to support their use in real-world reviews. More extensive evaluation is needed before advocating the widespread use of LLMs in systematic reviews.

Moreover, the LLM gave the wrong result in too many situations in this study for it to be considered 'safe' for full automation of any review task. Instead, it is likely to be more suited for a role where LLM predictions can be used to highlight likely relevant text and thus help reviewers to spot relevant information more rapidly. Still, a representative evaluation in the context of real systematic review projects is needed before making more recommendations about LLMs' value in systematic review automation.

## 10. Acknowledgements

Many thanks are due to the Evidence Synthesis Hackathon, without which this study would not have happened. We also wish to acknowledge with thanks the hospitality and exceptional room space from the NIHR Innovation Observatory at the University of Newcastle. Thanks to Emma Wilson for her contributions at the hackathon. Thanks also to Dr Alison O'Mara-Eves for useful feedback on the table and the contribution of *Template In A Box* ©.

## 11. Data Availability

Spreadsheets with the data used in the analysis are provided in our Git-Hub repository. This includes data not shown in the manuscript, such as full BLEU and ROUGE scores for the PICO study.

Programming code, for reproducing the figures and evaluation of the EBM-NLP data, is available via a GITHUB repository: https://github.com/L-ENA/ES-hackathon-GPT-evaluation

The EBM-NLP dataset and scripts to access it are available via this GITHUB repository: https://github.com/bepnye/EBM-NLP

## 12. Funding sources

LS was funded by the National Institute for Health and Care Research (NIHR) [HSRIC-2016-10009/Innovation Observatory]. The views expressed are those of the author(s) and not necessarily those of the NIHR or the Department of Health and Social Care.

## 13. Conflict of interest

None declared

# A. References for the test sets

## A.1. Animal studies

## A.2. Human clinical studies

Adams JB, Audhya T and Geis E ; Gehn E ; Fimbres V ; Pollard EL ; Mitchell J ; Ingram J ; Hellmers R ; Laake D ; Matthews JS ; Li K ; Naviaux JC ; Naviaux RK ; Adams RL ; Coleman DM ; Quig DW ;. (2018). Comprehensive Nutritional and Dietary Intervention for Autism Spectrum Disorder-A Randomized, Controlled 12-Month Trial.. *Nutrients*, 10(3), pp..

Araujo MS, Silva LGD and Pereira GMA ; Pinto NF ; Costa FM ; Moreira L ; Nunes DP ; Canan MGM ; Oliveira MHS ;. (2022). Mindfulness-based treatment for smoking cessation: a randomized controlled trial.. *Jornal brasileiro de pneumologia : publicacao 17oolean17 da Sociedade Brasileira de Pneumologia e Tisilogia*, 47(6), pp.e20210254.

Bos J, Staiger PK and Hayden MJ ; Hughes LK ; Youssef G ; Lawrence NS ;. (2019). A randomized controlled trial of inhibitory control training for smoking cessation and reduction.. *Journal of consulting and clinical psychology*, 87(9), pp.831-843.

Chai S C, Davis K and Zhang Z ; Zha L ; Kirschner K F;. (2019). Effects of Tart Cherry Juice on Biomarkers of Inflammation and Oxidative Stress in Older Adults. *Nutrients*, 11(2), pp..

Díaz-Silveira C, Alcover CM and Burgos F ; Marcos A ; Santed MA ;. (2020). Mindfulness versus Physical Exercise: Effects of Two Recovery Strategies on Mental Health, Stress and Immunoglobulin A during Lunch Breaks. A Randomized Controlled Trial.. *International journal of environmental research and public health*, 17(8), pp..

Dunsiger S, Emerson JA and Ussher M ; Marcus BH ; Miranda R Jr; Monti PM ; Williams DM ;. (2021). Exercise as a smoking cessation treatment for women: a randomized controlled trial.. *Journal of behavioral medicine*, 44(6), pp.794-802.

Enestvedt B K and Fennerty M B; Eisen G M;. (2011). Randomised clinical trial: MiraLAX vs. Golytely – a controlled study of efficacy and patient tolerability in bowel preparation for colonoscopy. *Aliment Pharmacol Ther*, 33(1), pp.33-40.

Kernc D, Strojnik V and Vengust R ;. (2018). Early initiation of a strength training based rehabilitation after lumbar spine fusion improves core muscle strength: a randomized controlled trial. *J Orthop Surg Res*, 13(1), pp.151.

Nixon A C, Bampouras T M; Gooch H J; Young H M. L; Finlayson K W; Pendleton N and Mitra S ; Brady M E; Dhaygude A P;. (2020). The EX-FRAIL CKD trial: a study protocol for a pilot randomised controlled trial of a home-based Exercise programme for pre-frail and FRAIL, older adults with Chronic Kidney Disease. *BMJ Open*, 10(6), pp.e035344.

Parekh D J, Reis I M; Castle E P; Gonzalgo M L; Woods M E; Svatek R S; Weizer A Z; Konety B R; Tollefson M and Krupski T L; Smith N D; Shabsigh A ; Barocas D A; Quek M L; Dash A ; Kibel A S; Shemanski L ; Pruthi R S; Montgomery J S; Weight C J; Sharp D S; Chang S S; Cookson M S; Gupta G N; Gorbonos A ; Uchio E M; Skinner E ; Venkatramani V ; Soodana-Prakash N ; Kendrick K ; Smith J A; Jr ; Thompson I M;. (2018). Robot-assisted radical cystectomy versus open radical cystectomy in patients with bladder cancer (RAZOR): an open-label, randomised, phase 3, non-inferiority trial. *Lancet*, 391(10139), pp.2525-2536.

**A.3. Human social science studies**

Berhanu Della, Okwaraji Yemisrach Behailu and Defar Atkure ; Bekele Abebe ; Lemango Ephrem Tekle; Medhanyie Araya Abrha; Wordofa Muluemebet Abera; Yitayal Mezgebu ; W

## B. Example of prompt development for animal studies

In the training set for animal studies, we first modified the original prompts using our prior understanding to increase their relevance to the animal literature. This development is summarised in Table 1.

Study designs are infrequently reported in the abstracts of such studies, so we simplified our prompt to request a description of study or experiment *type*. Responses to the prompt requesting subject details were often inconsistent in our earlier tests, so we modified the prompt to be more explicit about the types of information we wanted to extract. The Boolean prompts for animal and human study type didn't require further modification and seemed to already perform well in initial tests.

Study "arms" isn't terminology typically applied to animal studies, so we modified the comparisons prompt to "animal cohorts" instead and added that the interventions should also be detailed here. The extracted information still lacked detail, so we experimented by moving the order of the prompts so that the comparisons prompt was asked first. This markedly increased the amount of detailed information about each comparison extracted (e.g. including the animal model and dose of the drug). Asking GPT-4 to "list" these cohorts seemed to result in a more structured output, with separators between each animal group name.

The most challenging prompt development task was for outcome measures. In the training set, we iterated through approximately 20 different prompts to achieve full extraction of all outcome measures across both studies. In most scenarios, the prompt would generate a partial list of outcomes in one paper and a full list in the other paper. Changing the label from "outcomes" to "outcome_measure" and eventually "outcome_measures" led to significant improvements in results. Adjusting the prompt to request a "complete list" of outcomes led to a greater number of outcomes being reported in a sensible way. We also found it was important to not explicitly mention "comparison" here, perhaps because the abstract is not always so explicit in comparing groups of animals when reporting the main findings. Finally, the use of "biological parameter" in the prompt, while potentially limiting in scope, improved performance versus the non-specific phrase "outcomes".

**Table B.1: change from original 'generic' prompt to the one tailored for animal studies**

| Generic original prompt | Final Prompt used for assessment: Animal studies |
|---|---|
| study_design: string // Describe the study design. | study_design: string // Describe the type of study or experiment |
| Subjects: string //describe the subjects of this study | Subjects: string //describe the age, sex, and population characteristics of animals used in this study |
| isHuman: boolean // is this a study carried on human subjects? | isHuman: boolean // is this a study carried on human subjects? |
| isAnimal: boolean // is this a study carried on animal subjects? | isAnimal: boolean // is this a study carried on animal subjects? |
| subjects_number: number // how many subjects were used in this study? | subjects_number: number // how many subjects were used in this study? |
| Comparison_names: string // What are the names of the arms compared within this study? | Comparison_names: string // List animal cohorts involved in the study and any interventions they recieved |
| Outcomes: string // What were the main outcomes measured? | outcome_measures: string // A complete list of all specific biological parameters assessed for experimental groups |

## C. Summary of prompts used in the evaluation

| Category | Prompts: Human social science studies | Prompts: Animal studies | Prompts: Human clinical studies |
|---|---|---|---|
| Study design | study_design: string // Describe the study design. | Study_design: string // Describe the type of study or experiment | study_type: string // What is the research design of this paper? |
| Number of arms | arm_count: number // the number of arms in this trial | | |
| Causal inference method | causal_inference: string // Describe the causal inference method used to estimate intervention effectiveness in this study. | | |
| Country | study_country: string // the ISO Alpha-3 code of the country or countries where the study was conducted | | |
| Subjects | participants: string // give a full description of the participants in all groups of the study | Subjects: string //describe the age, sex, and population characteristics of animals used in this study | subjects: string // Description of the patients that participated in this study |
| Study on humans | isHuman: boolean // is this a study carried out with human participants? | isHuman: boolean // is this a study carried on human subjects? | isHuman: boolean // is this a study with human subjects? |
| Study on animals | isAnimal: boolean // is this a study carried out on animal subjects? | isAnimal: boolean // is this a study carried on animal subjects? | isAnimal: boolean // is this a study with animal subjects? |
| N of subjects in study | n_participants: number // total number of participants in all arms of the study | subjects_number: number // how many subjects were used in this study? | Number_subjects: number // how many subjects were used in this study? |
| Intervention description | intervention_descriptions: string // full and detailed description of the interventions that were compared within this study | | |
| Comparisons | comparisons: string // what treatments or conditions were compared with a control group, and which were compared with each other? | Comparison_names: string // List animal cohorts involved in the study and any interventions they recieved | Comparison_names: string // Describe the experimental arms in this trial and which interventions were given each |

| | | outcome_measures: string // A complete list of all specific biological parameters assessed for experimental groups | |
|---|---|---|---|
| Outcomes measured | Outcomes: string // What were the main outcomes measured? | | Outcomes: string // What were the main outcomes measured? |

### D. Example of a full JSON prompt

The box below shows the full (formatted) JSON prompt submitted to the GPT-4 API for the first article in the "Human Clinical Test Set" (Adams 2018) – prompts were not submitted with formatting, which we added to make the text more human-readable.

```json
{
    "messages": [
        {
            "role": "system",
            "content": "You extract data from the text provided below into a JSON object of the shape provided below. If the data is not in the text return 'false' for that field. \nShape: {study_type: string // What is the research design of this paper?,\nsubjects: string // Description of the patients that participated in this study,\nisHuman: boolean // is this a study with human subjects?,\nisAnimal: boolean // is this a study with animal subjects?,\nnumber_subjects: number // how many subjects were used in this study?,\nOutcomes: string // What were the main outcomes measured?,\nComparison_names: string // Describe the experimental arms in this trial and which interventions were given each,\n}"
        },
        {
            "role": "user",
            "content": "Text: This study involved a randomized, controlled, single-blind 12-month treatment study of a comprehensive nutritional and dietary intervention. Participants were 67 children and adults with autism spectrum disorder (ASD) ages 3-58 years from Arizona and 50 non-sibling neurotypical controls of similar age and gender. Treatment began with a special vitamin/mineral supplement, and additional treatments were added sequentially, including essential fatty acids, Epsom salt baths, carnitine, digestive enzymes, and a healthy gluten-free, casein-free, soy-free (HGCSF) diet. There was a significant improvement in nonverbal intellectual ability in the treatment group compared to the non-treatment group (+6.7 ± 11 IQ points vs. -0.6 ± 11 IQ points, p = 0.009) based on a blinded clinical assessment. Based on semi-blinded assessment, the treatment group, compared to the non-treatment group, had significantly greater improvement in autism symptoms and developmental age. The treatment group had significantly greater increases in EPA, DHA, carnitine, and vitamins A, B2, B5, B6, B12, folic acid, and Coenzyme Q10. The positive results of this study suggest that a comprehensive nutritional and dietary intervention is effective at improving nutritional status, non-verbal IQ, autism symptoms, and other symptoms in most individuals with ASD. Parents reported that the vitamin/mineral supplements, essential fatty acids, and HGCSF diet were the most beneficial."
        }
    ],
    "temperature": 0.0,
    "frequency_penalty": 0,
```

```
    "presence_penalty": 0,
    "top_p": 0.95
}
```

## E. Results for each study in the test set

| Short Title | Study domain | Study design | Number of arms | Causal inference method | Country | Subjects | Study on humans | Animal study | N subjects in study | Intervention description | Comparisons | Outcomes measured |
|---|---|---|---|---|---|---|---|---|---|---|---|---|
| Cheng (2015) | Animal | P | | | | √ | √ | √ | √ | | √ | √ |
| Gong (2022) | Animal | P | | | | √ | √ | √ | √ | | √ | P |
| Huang (2006) | Animal | P | | | | √ | √ | √ | √ | | √ | √ |
| Lee (2015) | Animal | X | | | | √ | √ | √ | √ | | √ | √ |
| Liu (2022) | Animal | P | | | | √ | √ | √ | √ | | √ | √ |
| Tauchi (2012) | Animal | √ | | | | P | √ | √ | √ | | √ | √ |
| Wang (2011) | Animal | X | | | | √ | √ | √ | √ | | P | P |
| Wang (2019) | Animal | √ | | | | √ | √ | √ | √ | | √ | √ |
| Wang (2023) | Animal | P | | | | √ | √ | √ | √ | | √ | P |
| Yang (2022) | Animal | P | | | | P | √ | √ | √ | | √ | √ |
| Adams (2018) | Human clinical | √ | | | | √ | √ | √ | √ | | √ | P |
| Araujo (2022) | Human clinical | √ | | | | √ | √ | √ | √ | | √ | √ |
| Bos (2019) | Human clinical | √ | | | | √ | √ | √ | √ | | √ | √ |
| Chai (2019) | Human clinical | √ | | | | √ | √ | √ | √ | | √ | X |
| Díaz-Silveira (2020) | Human clinical | √ | | | | √ | √ | √ | √ | | √ | √ |
| Dunsiger (2021) | Human clinical | √ | | | | P | √ | √ | √ | | P | P |
| Enestvedt (2011) | Human clinical | √ | | | | √ | √ | √ | √ | | √ | P |
| Kernc (2018) | Human clinical | √ | | | | √ | √ | √ | √ | | √ | √ |
| Nixon (2020) | Human clinical | P | | | | P | √ | √ | √ | | P | √ |
| Parekh (2018) | Human clinical | √ | | | | P | √ | √ | P | | √ | P |
| Berhanu (2020) | Human social science | X | √ | √ | √ | √ | √ | √ | X | P | P | √ |
| Buller (2016) | Human social science | P | √ | X | √ | P | √ | √ | √ | √ | √ | P |
| Charandabi (2010) | Human social science | √ | √ | X | √ | √ | √ | √ | X | √ | √ | √ |
| Cooper (2020) | Human social science | √ | √ | X | √ | √ | √ | √ | √ | √ | √ | √ |
| Czubak (2020) | Human social science | X | √ | √ | X | √ | X | √ | √ | √ | √ | √ |
| Eriksson (2014) | Human social science | X | √ | √ | √ | √ | √ | √ | √ | √ | √ | √ |
| Fu (2014) | Human social science | X | X | √ | √ | √ | X | √ | √ | P | X | √ |
| Giné (2009) | Human social science | √ | √ | √ | √ | √ | √ | √ | √ | √ | √ | P |
| Miranda (2016) | Human social science | X | √ | √ | √ | √ | X | √ | √ | P | P | √ |
| Pinto (2016) | Human social science | X | X | X | √ | X | √ | √ | √ | √ | X | √ |

## F. Overall aggregated results (N=30)

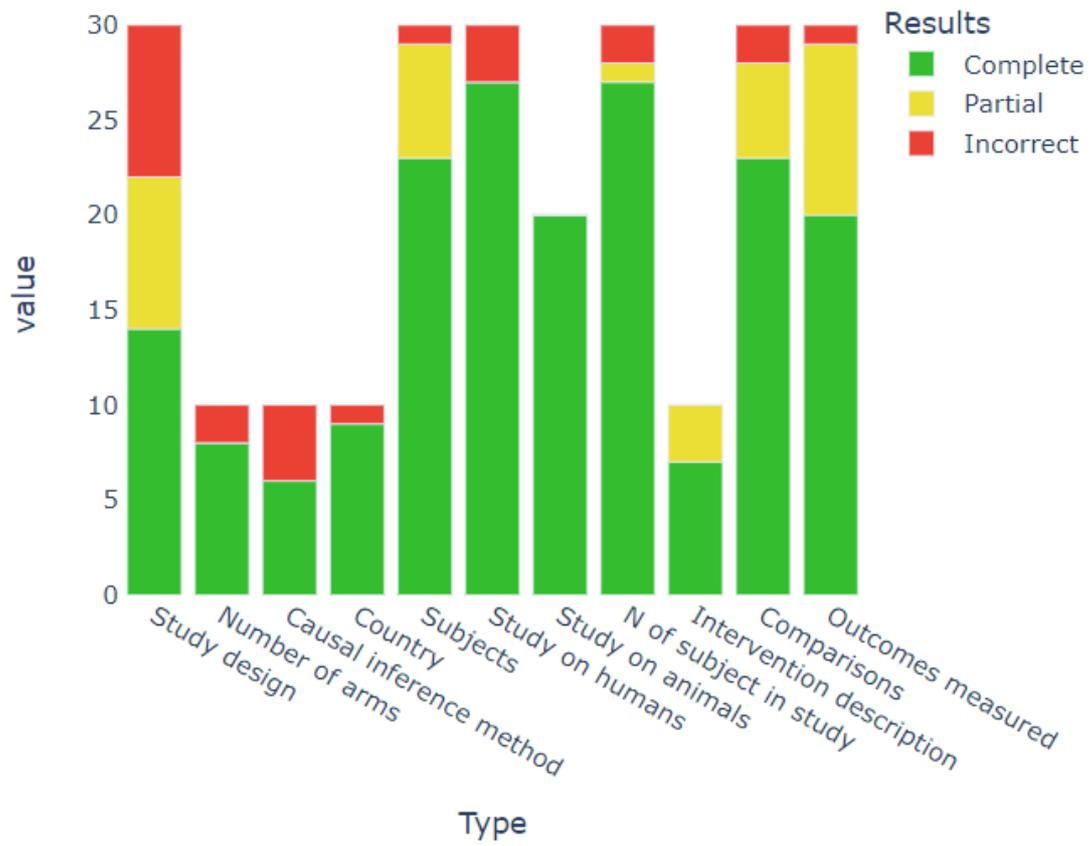

## G. Variability

The classification was applied by a single author, who had also been involved in the main evaluations (above) and according to the following criteria:

- "**NA**": for studies of types "clinical" and "animal" was applied to the questions that were unique for the social science domain.
- "**Equal**": answers provided by GPT-4 didn't change in the two rounds.
- "**Small change**": when the answers were different, but the meaning of the two answers did not change, for example, "RCT" vs. "Randomised controlled trial".
- "**Substantive change**": when the meaning of the answers was different, and different enough to likely change the original "Correct", "Incomplete", "Incorrect" classifications made in the core part of this evaluation.

The aggregated results are in Table G.1.

**Table G.1: variability between identical prompts**

| TOT | Equal | Small Change | Substantive Change |
|---|---|---|---|
| 250 | 204 | 35 | 11 |
| Percentage | 81.60% | 14.00% | 4.40% |

However, these figures fail to give the full picture, as each single question could differ in what kind of answer it admitted, distinguishing between "Boolean", "Number" and "String" answer type. Interestingly, we observed no variation in answers given to "Boolean" and "Number" questions. Table G.2 shows the updated variability measures, but calculated by considering only the answers of "String" type.

**Table G.2: variability between identical prompts, limited to "String" answers**

| TOT (strings) | Equal | Small Change | Substantive Change |
|---|---|---|---|
| 150 | 104 | 35 | 11 |
| Percentage | 69.33% | 23.33% | 7.33% |

These findings probably warrant further investigation, as it is possible that tweaking the "top_p" parameter (which was set to 0.95 throughout) might reduce variability to zero. Otherwise, variability itself might be used to identify answers for which the machine is more likely to produce mistakes, however attempting to determine if this is the case is beyond the scope of this evaluation.

| Short Title | Study design | Number of arms | Causal inf. method | Country | Subjects | Study on humans | Study on animals | N of subject in study | Intervention description | Comparisons | Outcomes measured |
|---|---|---|---|---|---|---|---|---|---|---|---|
| Cheng (2015) | Equal | NA | NA | NA | Equal | Equal | Equal | Equal | NA | Equal | Small change |
| Gong (2022) | Small change | NA | NA | NA | Equal | Equal | Equal | Equal | NA | Equal | Substantive Change |
| Huang (2006) | Equal | NA | NA | NA | Equal | Equal | Equal | Equal | NA | Equal | Equal |
| Lee (2015) | Equal | NA | NA | NA | Equal | Equal | Equal | Equal | NA | Equal | Equal |
| Liu (2022) | Small change | NA | NA | NA | Equal | Equal | Equal | Equal | NA | Equal | Small change |
| Tauchi (2012) | Small change | NA | NA | NA | Equal | Equal | Equal | Equal | NA | Equal | Equal |
| Wang (2011) | Equal | NA | NA | NA | Small change | Equal | Equal | Equal | NA | Small change | Equal |
| Wang (2019) | Small change | NA | NA | NA | Equal | Equal | Equal | Equal | NA | Equal | Small change |
| Wang (2023) | Equal | NA | NA | NA | Equal | Equal | Equal | Equal | NA | Equal | Equal |
| Yang (2022) | Equal | NA | NA | NA | Substantive Change | Equal | Equal | Equal | NA | Small change | Equal |
| Adams (2018) | Equal | NA | NA | NA | Equal | Equal | Equal | Equal | NA | Equal | Equal |
| Araujo (2022) | Equal | NA | NA | NA | Equal | Equal | Equal | Equal | NA | Small change | Equal |
| Bos (2019) | Equal | NA | NA | NA | Equal | Equal | Equal | Equal | NA | Small change | Equal |
| Chai (2019) | Equal | NA | NA | NA | Equal | Equal | Equal | Equal | NA | Small change | Substantive Change |
| Diaz-Silveira (2020) | Equal | NA | NA | NA | Equal | Equal | Equal | Equal | NA | Equal | Equal |

| Study | | | | | | | | | | | |
|---|---|---|---|---|---|---|---|---|---|---|---|
| Dunsiger (2021) | Small change | NA | NA | NA | Small change | Equal | Equal | Equal | NA | Substantive Change | Small change |
| Enestvedt (2011) | Small change | NA | NA | NA | Small change | Equal | Equal | Equal | NA | Substantive Change | Equal |
| Kernc (2018) | Equal | NA | NA | NA | Equal | Equal | Equal | Equal | NA | Substantive Change | Equal |
| Nixon (2020) | Small change | NA | NA | NA | Small change | Equal | Equal | Equal | NA | Substantive Change | Small change |
| Parekh (2018) | Equal | NA | NA | NA | Small change | Equal | Equal | Equal | NA | Small change | Equal |
| Berhanu (2020) | Equal | Equal | Small change | Equal | Equal | Equal | Equal | Equal | Equal | Equal | Small change |
| Buller (2016) | Equal | Equal | Equal | Equal | Equal | Equal | Equal | Equal | Equal | Small change | Equal |
| Charandabi (2010) | Equal | Equal | Equal | Equal | Equal | Equal | Equal | Equal | Equal | Small change | Equal |
| Cooper (2020) | Equal | Equal | Equal | Equal | Small change | Equal | Equal | Equal | Equal | Equal | Equal |
| Czubak (2020) | Equal | Equal | Equal | Substantive Change | Substantive Change | Equal | Equal | Equal | Equal | Equal | Small change |
| Eriksson (2014) | Equal | Equal | Equal | Equal | Equal | Equal | Equal | Equal | Equal | Equal | Equal |
| Fu (2014) | Equal | Equal | Equal | Equal | Equal | Equal | Equal | Equal | Substantive Change | Equal | Small change |
| Giné (2009) | Equal | Equal | Equal | Equal | Equal | Equal | Equal | Equal | Small change | Small change | Small change |
| Miranda (2016) | Equal | Equal | Small change | Equal | Equal | Equal | Equal | Equal | Substantive Change | Equal | Equal |
| Pinto (2016) | Equal | Equal | Equal | Equal | Equal | Equal | Equal | Equal | Equal | Equal | Small change |

# H. Template for future evaluations

Table H.1: issues and considerations

| Topic | Issues | What we did |
|---|---|---|
| Research question | Given the variability in responses from generative LLMs, it is important to investigate variability as well as accuracy | Evaluate the accuracy and variability of using a LLM for data (information) extraction |
| Dataset(s) | Data must reflect the variability in the domain of interest.<br><br>Data records used in evaluation must not have been used at all in prompt development | We selected records from three domains to explore variability; and also a widely-used dataset for PICO extraction<br><br>We split our data into train / test sets and did not look at the test set until we were evaluating LLM performance. |
| Prompt development | Researchers often iteratively design and test prompts, and sometimes use chains of prompts within the same 'conversation' with the LLM. All of this can affect performance and replicability, so needs to be described in detail. | We described above our approach to designing prompts, and specify precisely the prompts used. |
| How do you assess accuracy? | It can be difficult to assess accuracy of LLM output, as the same prompt can yield different outputs – but this output may, or may not, differ in meaning.<br><br>There are some automated approaches (e.g. BLEU and ROUGE scores), but these need to be validated and justified. High quality human assessment is helpful, though hard to obtain in high volume.<br><br>Accuracy can also be assessed using different metrics – e.g. % accuracy, or using different ordinal scales. | We created 'gold standard' human-agreed data and had two people assessing each record to increase reliability.<br><br>We also used human assessment in the PICO dataset, and tested the validity of automated approaches (BLEU and ROUGE scores), but found that they lacked validity in our use case, so preferred the human assessment.<br><br>Given that output can take different forms without changing essential meaning, we captured this by assessing output as 'correct', 'incorrect', or 'partially correct'. (i.e. output did not need to be identical if it was essentially correct) |
| Response stability | LLMs can give different output to the same prompt on repeated tests. Evaluations should assess the consequences of this behaviour. | We carried out a separate analysis where the LLM was repeatedly prompted with same prompts used in the primary analysis. |
| How are data analysed? | Many statistical and qualitative approaches might be used. Conventional significance tests may be appropriate with larger datasets, but qualitative assessment of accuracy often also plays a role. | Given this was an exploratory study with relatively small numbers of records we were conservative in our analytical approach and presented descriptive statistics. |
| How do you interpret the results? | Systematic reviews conventionally require a high degree of accuracy, as their results often affect decisions that affect people's lives. The critical question to ask when considering the use of a LLM for data extraction is whether its use might increase the risk that the review will generate wrong or misleading conclusions. | We found that, while the LLM was surprisingly accurate some of the time, it was also less accurate than our human assessors. We were therefore cautious in our conclusions, recommending that it was not yet ready for use in any 'full automation' task. |

**Box H.1: template for evaluating a LLM, developed with anticipated application to a systematic review task**

> **Research question/s:** Specify the question/s you are trying to answer with your evaluation.
> **Model parameter/algorithm:** Name the model you are planning to evaluate.
> **Comparator/s**: Name the algorithm/s or form/s of 'gold standard' that you are comparing to. Make clear whether they are an alternative model/algorithm or human generated data.
> **Performance measures:** Name the variable/s to be measured that are anticipated to be dependent on the model/algorithm.
> **Variables of interest:** State variables or conditions that might be related to the performance of the model/algorithm or the generalisability of the findings.
> **Dataset:** Specify how the data will be acquired. Name the dataset if using a pre-existing source. Explain any new data that needs to be generated to answer the research question/s.
> **Sub-task/s:** Note any distinct (possibly standalone) tasks that are required for enabling the evaluation, and whether the sub-tasks require their own self-contained evaluation.
> **Repeated trials/simulation runs:** State how many times will the experiment be performed. This can be used to assess stability of performance.
> **Analysis:** State how data will be analysed.

**Box 2: Populated template for evaluating an LLM for a systematic review task with two examples**

> **Research question/s:** Specify the question/s you are trying to answer with your evaluation.
>
> - *Comparison study: (1) How does the extraction of data items typical of those extracted in systematic reviews compare between a LLM and humans? (2) Does performance differ according to domain of question or domain of research? (3) How stable are the results from the LLM? (i.e. does repeated prompting yield the same responses?)*
> - *PICO study: (1) How does the extraction of data items typical of those extracted in systematic reviews compare between a LLM and humans?*
>
> **Model parameter/algorithm:** name the model you are planning to evaluate.
>
> - *Comparison study: LLM – GPT-4 with prompts created in EPPI Reviewer. The model used will be '2023-07-01-preview', accessed via the Azure OpenAI API on 13-15 December 2023.*
> - *PICO study: LLM – GPT-4 with prompts created in EPPI Reviewer. The model used will be '2023-07-01-preview', accessed via the Azure OpenAI API on 13-15 December 2023.*

**Comparator/s:** Name the algorithm/s or form/s of 'gold standard' that you are comparing to. Make clear whether they are an alternative model/algorithm or human generated data.

- *Comparison study: Human extracted text - new 'gold standard' human-agreed data.*
- *PICO study: Human extracted text - text that has been extracted by a diverse range of expert and non-expert humans.*

**Performance measures:** Name the variable/s to be measured that are anticipated to be dependent on the model/algorithm.

- *Comparison study: Primary outcome is accuracy (human assessment of whether the LLM extraction is complete, partial, or incorrect relative to the comparison). Outcome measure for stability of the model will be a human assessment of whether the response to an identical prompt is equal, represents a small change, or represents a substantive change.*
- *PICO study: Primary outcome is accuracy (human assessment of whether the LLM extraction is complete, partial, or incorrect relative to the comparison). Assess the applicability of automated metrics BLEU and ROUGE scales.*

**Variables of interest:** State variables or conditions that might be related to the performance of the model/algorithm or the generalisability of the findings.

- *Comparison study: Study type (animal, human clinical, human social science). Type of data extracted (study design, number of arms, causal inference method, country, subjects, study on humans, study on animals, N of subject in study, intervention description, comparisons, outcomes measured).*
- *PICO study: Type of data extracted (participants, intervention/control, and outcomes).*

**Dataset:** Specify how the data will be acquired. Name the dataset if using a pre-existing source. Explain any new data that needs to be generated to answer the research question/s.

- *Comparison study: novel dataset of 36 cases built from 12 human social science, 12 animal, and 12 human clinical studies. In each of the three domains, split the subset into two studies in the 'train' set (to be used for prompt development) and 10 in the 'test' set to be used for testing the automated data extraction against). A 'gold standard' human data extraction will be generated for each study. Human decisions on the accuracy of LLM versus EBM-NLP extractions need to be manually curated. See also sub-task: prompt development.*

- *PICO study:* 100 studies from the EBM-NLP dataset. Human decisions on the accuracy *of LLM versus EBM-NLP extractions need to be manually curated. See also sub-task: prompt development.*

**Sub-task/s:** Note any distinct (possibly standalone) tasks that are required for enabling the evaluation, and whether the sub-tasks require their own self-contained evaluation.

- *Comparison study: Prompt development required to ensure viable prompts to submit to GPT-4. Test for prompt sensitivity (e.g., different composition or structure).*
- *PICO study: Prompt development required to ensure viable prompts to submit to GPT-4. Test for prompt sensitivity (e.g., different composition or structure).*

**Repeated trials/simulation runs:** State how many times will the experiment be performed. This can be used to assess stability of performance.

- *Comparison study: two trials. Repeat the automated data extraction a second time against the test set using identical prompts to the first round.*
- *PICO study: one trial*

**Analysis:** State how data will be analysed.

- *Comparison study: Accuracy: descriptive statistics (percentages) for overall accuracy and by type of data extracted (study design, number of arms, causal inference method, country, subjects, study on humans, study on animals, N of subject in study, intervention description, comparisons, outcomes measured). Also, separate results for three subsets of study type (animal, human clinical, human social science). Stability: descriptive statistics (percentages) of assessments (equal, small change, substantive change) for identical prompts and for identical prompts limited to 'string' answers.*
- *PICO study: Accuracy: descriptive statistics for overall accuracy and by type of data extracted (participants, intervention/control, and outcomes). Use of BLEU / ROUGE scales if applicable.*